\documentclass[wcp]{jmlr}

\makeatletter
\renewcommand*{\thanks}[1]{%
  \footnotemark
  \protected@xdef\@thanks{\@thanks
    \protect\footnotetext[\arabic{footnote}]{#1}}%
}
\makeatother

\usepackage{longtable}
\usepackage{booktabs}
\usepackage{lineno}
\usepackage{appendix}
\usepackage{caption}
\usepackage{subcaption}
\usepackage{algorithm2e}

\makeatletter
\let\Ginclude@graphics\@org@Ginclude@graphics 
\makeatother

\jmlrvolume{222}
\jmlryear{2023}
\jmlrworkshop{ACML 2023}

\RestyleAlgo{ruled}

\title[Rate Card Transformer]{Unveiling the Power of Self-Attention for Shipping Cost Prediction: The Rate Card Transformer}

\author{\Name{P Aditya Sreekar}\Email{sreekarp@amazon.com}\thanks{These authors contributed equally to this work.}\\
        \addr Amazon
        \AND
        \Name{Sahil Verma}\Email{vrsahil@amazon.com}\footnotemark[1]\\
        \addr Amazon
        \AND
        \Name{Varun Madhavan}\Email{varunmadhavan@iitkgp.ac.in}\thanks{Work done during internship at Amazon}\\
        \addr Indian Institute of Technology, Kharagpur
        \AND
        \Name{Abhishek Persad}\Email{persadap@amazon.com}\\
        \addr Amazon
        }

\editors{Berrin Yan{\i}ko\u{g}lu and Wray Buntine}
\begin{document}

\maketitle

\begin{abstract}
Amazon ships billions of packages to its customers annually within the United States. Shipping cost of these packages are used on the day of shipping (day 0) to estimate profitability of sales. Downstream systems utilize these days 0 profitability estimates to make financial decisions, such as pricing strategies and delisting loss-making products. However, obtaining accurate shipping cost estimates on day 0 is complex for reasons like delay in carrier invoicing or fixed cost components getting recorded at monthly cadence. Inaccurate shipping cost estimates can lead to bad decision, such as pricing items too low or high, or promoting the wrong product to the customers. Current solutions for estimating shipping costs on day 0 rely on tree-based models that require extensive manual engineering efforts. In this study, we propose a novel architecture called the Rate Card Transformer (RCT) that uses self-attention to encode all package shipping information such as package attributes, carrier information and route plan. Unlike other transformer-based tabular models, RCT has the ability to encode a variable list of one-to-many relations of a shipment, allowing it to capture more information about a shipment. For example, RCT can encode properties of all products in a package. Our results demonstrate that cost predictions made by the RCT have 28.82\% less error compared to tree-based GBDT model. Moreover, the RCT outperforms the state-of-the-art transformer-based tabular model, FTTransformer, by 6.08\%. We also illustrate that the RCT learns a generalized manifold of the rate card that can improve the performance of tree-based models.
\end{abstract}

\section{Introduction}
\label{sec:intro}

Amazon ships packages in the order of billions annually to its customers in the United States alone. The route planning for these packages is done on the day of shipping, \textit{day 0}. As part of this plan, the shipping cost for each package is estimated by breaking down the package journey into smaller legs, and calculating the cost of each leg using a rate card. Day 0 cost estimates are used to compute initial profitability estimates for accounting purposes, e.g.the estimate of profit/loss for each item as a result of a specific sale to a customer. These profitability estimates are used by several downstream services for decision making and planning.

However, the day 0 estimates may differ from the actual cost due to factors like improper rate-card configuration, incorrect package dimensions, wrong delivery address, etc. 
% In 2022, the total absolute difference between estimated and actual cost was estimated to be 25\% of the total shipping costs.
Inaccurate cost estimates cause skewed profitability estimates, which in turn leads to suboptimal financial decisions by downstream systems. For example, if the shipping cost of an item is consistently overestimated, then the item could be removed from the catalog. On the other hand, underestimated cost can lead pricing systems to lower the price of the item, leading to losses. Further, inaccurate estimation also leads us to promote wrong products to the customer, causing bad customer experience. To improve these shipping cost estimates, we propose a Transformer based deep learning model that accurately predicts the shipping cost at day 0.

In the context of shipping, a package is characterized by its physical dimensions, weight, and contents. It also include details about the carrier responsible for transporting it and the intended route. Additionally, a package is associated with a variable number of attributes that describe the item(s) inside and the various charges related to its shipment. Collectively, we refer to these attributes as the \textit{rate card} associated with the package. For tabular datasets like package rate cards, tree based models like Gradient Boosted Decision Trees (GBDT), XGBoost \citep{chen2016xgboost}, etc., are considered as state-of-the-art (SOTA) models. However, their effectiveness heavily relies on high-quality input features \citep{arik2019tabnet} which can require extensive feature engineering. For our use case, this problem is further accentuated by the fact that the target concept depends on high order combinatorial interactions between rate card attributes. For example, if the rate card is improperly configured for large containers with flammable substances  shipped from Washington DC to New York by ABC carrier, then the model has to learn to associate property combination $<size=large,\ item=flammable,\ source=Washington,\ destination=New\ York,\ carrier=ABC>$ with high deviation between estimated and actual costs. When dealing with feature combinations, considering all possible higher-order interactions between package properties may be impractical due to the exponential increase in the number of interactions with each increase in order, leading to the curse of dimensionality \citep{bishop2006pattern}. Another shortcoming of tree based models is their inability to handle a variable length list of features. A package may contain multiple items, and its ship cost can be broken down into multiple charges types. Previous experiments demonstrated that adding features engineered from multiple items and charges improved GBDT's performance. However, due to inability of tree based models to handle variable list of features, complete information from them could not be learned. 

In this paper, inspired by the recent success of transformers in tabular domain \citep{huang2020tabtransformer,somepalli2021saint,gorishniy2021revisiting}, we propose a novel architecture called the Rate Card Transformer (RCT) to predict ship cost on day 0. The proposed model is specifically designed to learn an embedding of rate card associated with a package. The RCT leverages self-attention mechanisms to effectively capture the interdependencies between various components in a rate card by learning interactions between input features. Specifically, our contributions in this work include:
\begin{itemize}
    \item Propose a novel architecture, \textit{Rate Card Transformer} (RCT), which leverages transformer architecture to learn a manifold of the rate card, to predict shipping cost on Day 0. Further, it is demonstrated that RCT outperforms both GBDTs and the state-of-the-art tabular transformer, FT-Transformer, \citep{gorishniy2021revisiting} in shipping cost prediction.
    \item Extensive experiments are performed to show that the learned embeddings are a sufficient representation of the rate card manifold, and self-attention layers are effective feature interaction learners. Ablation studies are performed to analyze the impact of number of transformer layers and attention heads on model performance. 
\end{itemize}

\section{Related Works}

Tree-based algorithms are widely used in machine learning for tabular data. Decision trees recursively split the data into multiple parts based on axis-aligned hyper-planes \citep{hastie2009elements}. Random Forests (RF) \citep{breiman2001random} and Gradient Boosted Decision Trees (GBDT) \citep{friedman2001greedy} are the most commonly used tree based ensembles. RF fits multiple decision trees on random subsets of the data and averages/polls the predictions to alleviate the overfitting characteristic of decision trees. GBDT, XGBoost \citep{chen2016xgboost}, and CatBoost \citep{prokhorenkova2018catboost} are boosted ensemble models that sequentially build decision trees to correct errors made by previous trees, leading to improved performance on complex datasets with non-linear relations. 

Recently, there has been a lot of interest in deep learning models for tabular data. Some methods introduce differentiable approximations of decision functions used in decision trees to make them differentiable \citep{hazimeh2020tree, popov2019neural}. These methods outperform pure tree based problem for some problem statements, however, they are not consistently better \citep{gorishniy2021revisiting}. Other methods have used attention mechanisms to adapt DL methods to tabular data \citep{arik2019tabnet, huang2020tabtransformer, gorishniy2021revisiting, somepalli2021saint, chen2022danets}. TabNet \citep{arik2019tabnet} proposes a sparse attention mechanism that is stacked in multiple layers to mimic the recursive splitting of decision trees. Inspired from the success of self-attention transformers \citep{vaswani2017attention} in many domains \citep{devlin2019bert, dosovitskiy2020vit, gong21b_interspeech} methods like TabTransformer \citep{huang2020tabtransformer}, FT-Transformer \citep{gorishniy2021revisiting} and SAINT \citep{somepalli2021saint} were proposed. TabTransformer embeds all categorical variables into a unified embedding space, and a sentence of categorical embeddings is passed through self-attention transformer layers. FT-Transformer further extends this by attending to numerical features as well, by using continuous embedding. SAINT builds on FT-Transformer by proposing a new kind of attention which captures interactions between samples of a batch. However, SAINT does not provide any advantage over FT-Transformer for our problem statement, because intersample attention is only effective when the number of dimensions is higher in comparision to the number of samples, thus we do not compare RCT against SAINT \citep{somepalli2021saint}.

\section{Methodology}

\begin{figure}
    \begin{subfigure}{0.72\textwidth}
      \centering
      \includegraphics[width=\textwidth]{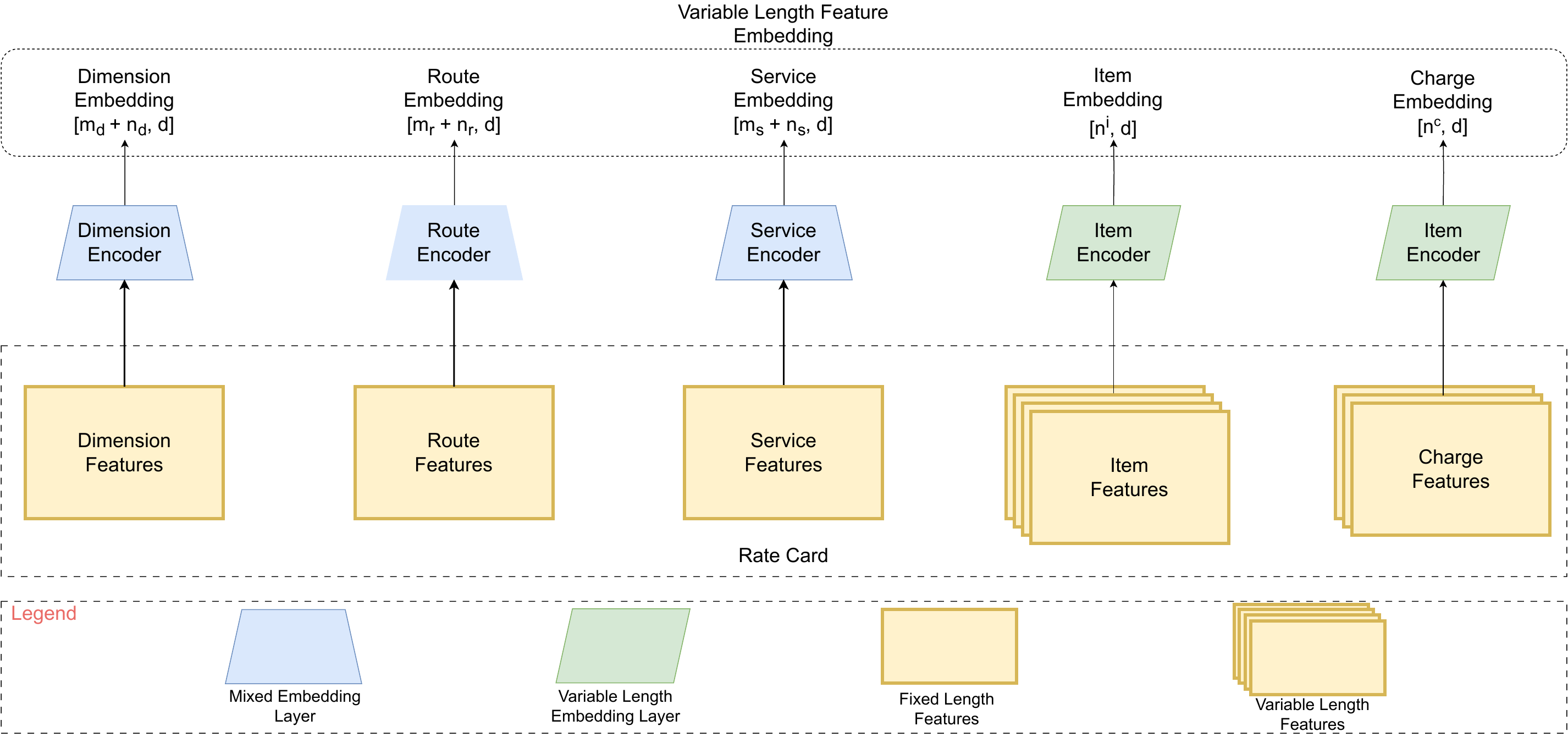}
      \caption{}
      \label{fig:rate-card-example}
    \end{subfigure}
    \begin{subfigure}{.27\textwidth}
      \centering
      \includegraphics[width=\linewidth]{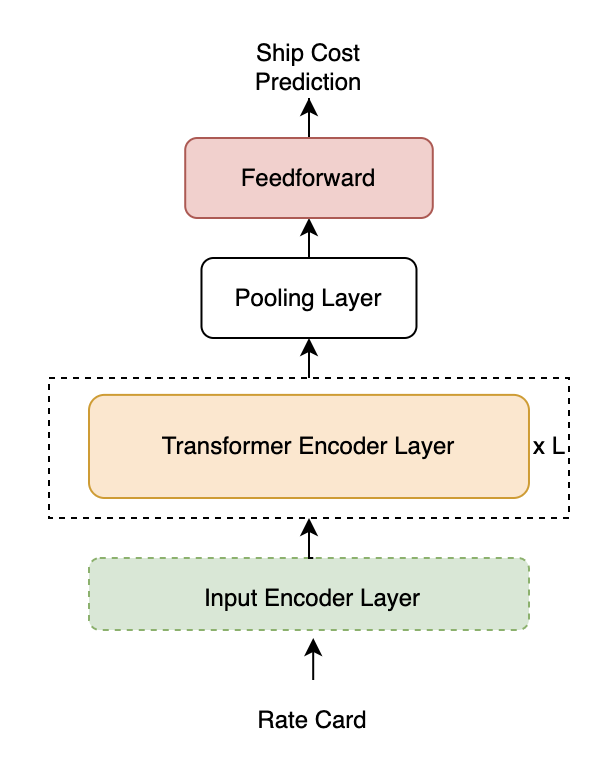}
      \caption{}
      \label{fig:rct_full}
    \end{subfigure}
    \caption{(\subref{fig:rate-card-example}) Input encoder layer of Rate Card Transformer. (\subref{fig:rct_full}) RCT Architecture}
    \label{fig:my_label}
\end{figure}

\subsection{Problem Statement}
\label{subsec:problem_statement}

Rate card information of a shipment is a collection of feature types - dimension, route, service, item, and charge, as illustrated in Fig. \ref{fig:rate-card-example}. Each type contains $m$ numerical and $n$ categorical features, represented as $\mathbf{x} \in \mathcal{S}[m,n]$. Formally, the rate card representation of a package is  $\mathbf{P} = \{\mathbf{d}, \mathbf{r}, \mathbf{s}, \{\mathbf{i}_{k}\}_{k=1}^{n^i}, \{\mathbf{c}_{k}\}_{k=1}^{n^c}\}$ where $\mathbf{d} \in \mathcal{S}[m_d,n_d]$ are dimensional features, $\mathbf{r} \in \mathcal{S}[m_r,n_r]$ are route features, $\mathbf{s} \in \mathcal{S}[m_s,n_s]$ are service features,  $\mathbf{i}_{k} \in \mathcal{S}[m_i,n_i]$ are features of $k^{th}$ item, $\mathbf{c}_{k} \in \mathcal{S}[m_c,n_c]$ are features of $k^{th}$ charge, and $n^i$, $n^c$ are the number of items and charges in the package. 

The objective is to make an estimate $\hat{C}$ of the unknown actual shipping cost $C$, given the \textit{day 0} heuristic estimate $C^A$ and the rate card $\mathbf{P}$. A functional mapping $f(\mathbf{P}, C^A; \theta) \approx \hat C$ with parameters $\theta$ is learned from a dataset $\mathcal{D} = \{\mathbf{P}_j, C_j, C^A_j\}_{j=1}^N$, of $N$ packages shipped in the past.

\subsection{Background}
\label{subsec:transformers}

The Transformer architecture \citep{vaswani2017attention} is constructed by stacking multiple encoder blocks, where each block takes a sequence of embeddings as input and outputs a sequence of context aware embeddings. The encoder block consists of a multi-head self-attention (MHSA) layer followed by a position-wise feed-forward layer, with residual connections and layer norm before each layer. The MHSA layer comprises multiple self-attention units called heads, which learn interactions between input embeddings. 

%vrsahil edit
The encoder layers are powerful feature aggregators when the input sequence consists of features. The encoder leverages MHSA layers to produce an interaction-aware representation of the input features. This is accomplished through the use of self-attention heads, which computes a weighted sum of the input feature embeddings. To determine the summation weights, called attention scores, the attention mechanism projects the input features using learned matrices into three subspaces, query $q_i$, key $k_i$ and value $v_i$. The attention score, between two features are computed using Eq.\ref{eq:attention}.

The attentions score $a_{i,j}$ quantifies the interaction between features $i$ and $j$. The raw attention scores are softmax normalized to ensure that they sum up to $1$. The interaction-aware representation of feature $i$, $o_i$, which considers interaction between all features, is computed using the equation Eq. \ref{eq:value-vector}.

\begin{gather}
    a_{i, j}(q_i, k_j) = softmax(\frac{q_i \cdot k_j}{\sqrt{d}}) \label{eq:attention} \\
    o_i = \sum_{j} a_{i, j} v_j \label{eq:value-vector}
\end{gather}

%vrsahil edit
The output sequence is then recursively passed through subsequent encoder layers, allowing each successive layer to learn higher order feature interactions. The transformer's depth controls the complexity of the learned representation, as deeper layers capture more complex interactions between features. Further, multiple self-attention heads are used in MHSA, enabling each head to attend to different feature sub-spaces and learning interactions between them,
cumulatively learning multiple independent sets of feature interactions.

\subsection{Rate Card Transformer}

The rate card of a package consists of multiple features types, namely dimensional, route, service, item, and charge (Fig. \ref{fig:rate-card-example}), where each feature type comprises multiple numerical and categorical features. The dimensional, route and service features are referred to as \textit{fixed length feature} types, because each of them have a fixed number of features. Fixed length feature types are embedded to a sequence of tokens using a \textit{mixed embedding layer} (MEL). For example, dimensional features $\mathbf{d} \in \mathcal{S}[m_d,n_d]$ are embedded to a $d$-dimensional token sequence of length $m_d + n_d$. The MEL contains multiple embedding blocks, one for each feature in the feature type being embedded. Embedding lookup tables are used for embedding categorical features, while numerical features are embedded using continuous embedding blocks, as introduced in \citep{gorishniy2021revisiting}. 

Unlike fixed length features, a package is associated with variable number of items and charges, thus these are referred to as variable length features. Each item is first embedded through a mixed embedding layer, creating a sequence of tokens. The sequence is reduced to a single token, creating a single embedding for each item. These set of layers is collectively referred to as \textit{variable length embedding layer}. Charges are also embedded similarly. After all features have been embedded, a token sequence of length $[(m_d + n_d) + (m_r + n_r) + (m_i + n_i) + n^i + n^c]$ is constructed. This is called the \textit{rate card embedding}.

The sequence of feature tokens is passed as input to a stack of $L$ Transformer encoder layers that are able to learn complex, higher order interactions between the features. Finally, the pooled Transformer output is fed to a feedforward layer to predict the shipping cost $\hat C$ as shown in Fig. \ref{fig:rct_full}. 

We call the complete architecture the \textit{Rate Card Transformer} (RCT). Trained to minimize the L1 loss between the predicted and actual shipping cost (Equation \ref{eq:loss}), RCT learns an effective representation of the dynamic rate card that allows it to accurately predict the shipping cost. 

\begin{equation}
    \label{eq:loss}
    \underset{\theta}{\mathrm{argmin}} \underset{\mathbf{P},C,C^A \sim \mathcal{D}}{\mathbb{E}}\left[\left|C-f\left(\mathbf{P},C^A\right)\right|\right]
\end{equation}

\section{Experiments}
In this section, the performance of the RCT is demonstrated on a dataset of packages shipped in 2022. The mean absolute error (MAE) between the predicted and actual shipping cost is selected as the performance metric, as it is representative of the absolute error in monetary terms. In this paper, the MAE values are normalized by the MAE of day 0 heuristic estimate, which is expressed as MAE percentage ($MAE_{\%}$). This metric emphasizes the improvement achieved against the heuristic baseline.

\begin{equation}
    \label{eq:mae-percentage}
    MAE_{\%} = \frac{\sum_i^N |C_i - \hat{C}_i|}{\sum_i^N |C_i - C^A_i|} \times 100
\end{equation}

\subsection{Experimental Setup}
\label{subsec:exp_setup}

\subsubsection{Architecture and Hyperameters}
\label{sec:hyperparams}
The embedding dimension was set to 128, and 6 transformer encoder layers were used, each with 16 self-attention heads. Adam optimizer \citep{kingma2014adam} with a starting learning rate of 0.0001 and a batch size of 2048 was used. To improve convergence, the learning rate was reduced by a factor of 0.7 every time the validation metric plateaued. The model code was implemented using the PyTorch \citep{prokhorenkova2018catboost} and PyTorch Lightning \citep{falcon2019pytorchlightning} frameworks.

\subsubsection{Data Preparation}
A training dataset of 10M packages was sampled from packages shipped during a 45-day period in 2022. The data was preprocessed by label encoding categorical features and standardizing numerical features. The test dataset contains all packages (without sampling) that were shipped during a separate, non-overlapping week from 2022.  

\subsubsection{Benchmark Methods}
\label{subsubsec:baseline}
We compare the performance of RCT against various models with increasing level of complexity: GBDT, AWS AutoGluon \citep{erickson2020autogluontabular}, Feedforward neural network, TabTransformer and FT-Transformer. For GBDT model, numerical features were not standardized, and target encoding \citep{daniele2001targetencoding} was used to encode categorical features instead of label encoding. AWS AutoGluon was configured to learn an ensemble of LightGBM \citep{ke2017lightgbm} models. A feedforward neural network containing 5 layers was used, the input to which was generated by embedding and concatenating dimension, route and service features. Publicly available implementations \footnote{\href{https://github.com/lucidrains/tab-transformer-pytorch}{https://github.com/lucidrains/tab-transformer-pytorch}} of TabTransformer and FT-Transformer were used, and all hyperparameters were made consistent with RCT. Since the baselines do not handle collections of items and charges, we only used dimension, route and service features.

\subsection{Baseline Comparisons}
\label{subsec:baseline_comparision}
Table \ref{tab:mae_results} compares RCT against the baseline models discussed in section \ref{subsubsec:baseline}. The models in the table are organized in increasing order of model complexity. Both tree based models, GBDT and AutoGluon, are performing at a similar level. Deep learning models consistently outperform tree based models, indicating that the proposed architecture is efficient for shipping cost prediction. Transformer based models have lower $MAE_\%$ scores than feedforward neural network, showing that transformers learn effective interaction. The RCT model outperforms both transformer models - TabTransformer and FT-Transformer (SOTA), suggesting that a custom architecture which encodes the latent structure of the rate card is contributing to the improved performance. Table \ref{tab:different-size-comp} compares the performance of FT-Transformer and RCT models at different model sizes. The results show that RCT outperforms FT-Transformer across all tested model sizes, showing indicates that encoding rate card structure provides performance benefits across varying model capacities.

\begin{table}
    \caption{(a) compares the performance of the RCT against various benchmarks, (b) compares the performance of GBDT baseline with GBDT trained with RCT embeddings. $MAE_{\%}$ is calculated as shown in Equation \ref{eq:mae-percentage}.}
    \begin{subfigure}{.35\textwidth}
        \centering
        \caption{}
        \label{tab:mae_results}
        \begin{tabular}{@{}cc@{}}
                \toprule
                \textbf{Model}             & \textbf{$MAE_{\%}$} \\ 
                \midrule
                % Day 0 estimate         & 100 \\ % 24.10             
                GBDT baseline          & 78.29 \\ % 18.87             
                AutoGluon              & 77.59 \\ % 18.70              
                Feed Forward           & 67.13 \\ % 16.18            
                TabTransformer         & 69.58 \\ % 16.77    
                FT-Transformer         & 59.33 \\ % 14.30             
                \textbf{RCT}           & \textbf{55.72} \\ % 13.43
                \bottomrule
            \end{tabular}
    \end{subfigure}
    \hfill
    \begin{subfigure}{.60\textwidth}
        \centering
        
        \caption{}
        \label{tab:rep_learned}
        \begin{tabular}{@{}cc@{}}
                 \toprule
                 \textbf{Model}                 & \textbf{$MAE_{\%}$} \\
                 \midrule
                 &\\
                 GBDT + manual features & 78.29 \\ % 18.87 
                 &\\
                 GBDT + manual features + RCT embedding & 68.50 \\ % 16.51 
                 &\\
                 GBDT + RCT embedding & 69.21 \\ % 16.68 
                 &\\
                 \bottomrule
            \end{tabular}
    \end{subfigure}
\end{table}

\begin{table}
\caption{\textbf{\textit{$MAE_\%$ comparison between RCT and FT-Transformer (SOTA for self-attention models)}}}
    \centering
    \begin{tabular}{ccccc}
    \toprule
        \textbf{Layers} & \textbf{nheads} & \textbf{d\_model} & \textbf{RCT $MAE_\%$} & \textbf{FT-Transformer $MAE_\%$} \\ \hline
        1 & 4 & 32 & 73.82\% & 76.31\% \\
        3 & 8 & 64 & 61.95\% & 63.11\% \\
        6 & 16 & 128 & 55.73\% & 59.34\% \\
    \bottomrule
    \end{tabular}
    \label{tab:different-size-comp}
\end{table}

\subsection{Does RCT learn effective representation of rate cards?}
\label{subsec:learned-rep-downstream}

%vrsahil_edit
Transformers have been shown to have strong representation learning capabilities in a variety of tasks. In this experiment, we investigate the
effectiveness of rate card representation learned by RCT. To evaluate this, we compare the performance of our GBT model with and without the learned rate card representation as an input feature.

The pooled output of the final Transformer layer is treated as the learned representation of the rate card. Adding this feature improved the performance of the GBDT by 9.79\% (refer Table \ref{tab:rep_learned}). Further, it was observed that even when all manually engineered features are dropped, the GBDT still performs comparably, with an MAE percentage of 69.21\%. This indicates that the learned representations of rate cards are not only effective at capturing better feature information, but are also sufficient representation of the package rate card. However, even with this feature, the GBDT has a ~13.5\% higher $MAE_\%$ than the RCT. This is likely because the RCT is trained end-to-end, while the GBDT uses features learned as part of a separate model. 

\begin{figure}[t]
    \centering
    \begin{subfigure}{0.4175\textwidth}
        \centering
        \includegraphics[width=\textwidth]{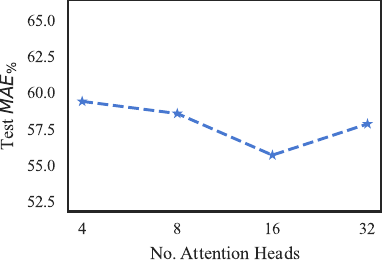}
        \caption{}
        \label{subfig:nheads_graph}
    \end{subfigure}
    \hfill
    \begin{subfigure}{0.4825\textwidth}
        \centering
        \includegraphics[width=\textwidth]{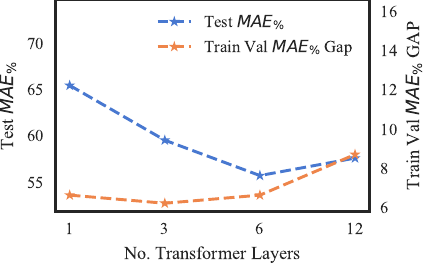}
        \caption{}
        \label{subfig:layers_graph}
    \end{subfigure}
    \caption{Figure \subref{subfig:nheads_graph} plots test $MAE_{\%}$ vs number of attention heads. Figure \subref{subfig:layers_graph} plots test $MAE_\%$ and train-val $MAE_\%$ gap vs number of transformer layers. $MAE_{\%}$ is calculated as shown in Equation \ref{eq:mae-percentage}.}
    \label{fig:my_label}
\end{figure}

\subsection{Does self-attention learn better interactions than feed forward neural networks? }
\label{subsec:package-feats}

In section \ref{subsec:baseline_comparision}, it was observed that feed forward (FF) neural networks were outperformed by transformers, leading to the hypothesis that self-attention is a superior interaction learner. This section aims to explore this hypothesis further by utilizing FF instead of self-attention to encode dimension, route and service features while limiting the width of self-attention to only the item and charge features. The output encodings of both FF and self-attention are concatenated and fed into an FF layer to predict shipping cost. As the self-attention width is decreased, it fails to capture the interactions between all rate card features. The resulting model exhibits a higher $MAE_\%$ of 64.73\% in comparison to the RCT’s 55.72\%. These results suggest that FF models are inferior interaction learners in comparison to transformers.

\begin{algorithm2e}[t]
    \caption{Extraction of most attended feature per head}\label{alg:interaction}
    \KwData{Package Dataset $\mathcal{D}$, self-attention heads $\mathcal{H}=\{\mathrm{h}_i\}$ and $N$ features}
    \KwResult{Heat map of most attended features}
    $heat\_map \gets []$\;
    \For{$\mathrm{h} \in \mathcal{H}$}{
        $attention\_importance \gets [0]_{i=0}^N$ \;
        \For{$\mathbf{P} \in \mathcal{D}$}{
            Compute attention map $A = \{a_{i,j}\}$ of $\mathrm{h}$ for $\mathbf{P}$\;
            Increment $attention\_importance[j]$ by $1$ for top five interaction $\{i,j\}$ from $A$\;
        }
        $attention\_importance \gets \frac{attention\_importance - min(attention\_importance)}{max(attention\_importance) - min(attention\_importance)}$
        Append $attention\_importance$ to $heat\_map$
    }
\end{algorithm2e}

\begin{figure}[t]
    \centering
    \begin{subfigure}{0.65\textwidth}
        \centering
        \includegraphics[width=\textwidth]{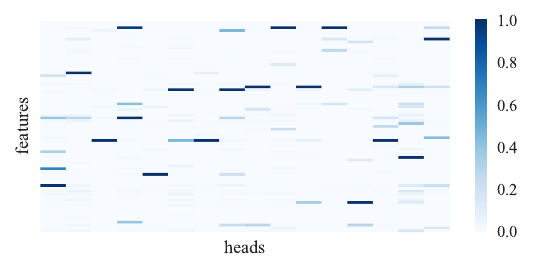}
        \caption{16 Attention Heads}
        \label{subfig:interaction1}
    \end{subfigure}
    % \hfill
    \begin{subfigure}{0.34\textwidth}
        \centering
        \includegraphics[width=\textwidth]{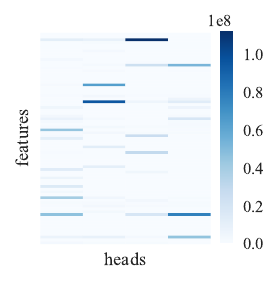}
        \caption{4 Attention Heads}
        \label{subfig:interaction2}
    \end{subfigure}
    \caption{Heatmaps generated from \ref{alg:interaction}. Each column shows the relative importance of each feature in a head, and each column corresponds to a different head.}
    \label{fig:interaction}
    % \vspace{-0.8cm}
\end{figure}

\subsection{Analysis of Self-Attention}
\label{subsec:feature-interactions}
In section \ref{subsec:transformers}, we discussed the proficiency of transformers in feature aggregation, owing to self-attention. In this section, ablation experiments are conducted to analyze the effect of attention depth and attention head count. Increasing the number of attention heads allows the model to learn more independent feature interactions. For this experiment, the model capacity is fixed at 128 dimensions, so an increase in the number of heads also reduces the complexity of interactions learned per head. Thus, choosing optimal head count is a trade-off between learning independent interactions and the complexity of each learned interaction. The trade-off can be observed in Fig. \ref{subfig:nheads_graph}, where the performance improves from 4 heads to 16 heads because the attention learned by each head is complex enough. However, the performance degrades when attention heads are increased from 16 to 32 because the complexity of heads has reduced substantially, negating the benefit of learning more independent interactions.

Next, we illustrate the effect of increasing the attention depth by adding transformer encoder layers. Deeper transformer networks learn more complex higher-order interactions, thereby enhancing the model's performance, as observed in Fig. \ref{subfig:layers_graph}. However, increasing the number of layers from 6 to 12 reduces the model's performance due to overfitting, caused by the rise in learnable parameter count. The evidence for overfitting can be found in Fig. \ref{subfig:layers_graph}, where the gap between train and val MAE increases by 30\% when moving from 6 to 12 layers. 

Finally, in Fig. \ref{fig:interaction}, we display the heat maps generated using Algorithm \ref{alg:interaction}. These heat maps illustrate the number of times each feature was attended to as part of the top five most attended features. Each column corresponds to a head, and each row corresponds to a feature. The heat map on the left was generated using RCT with $nheads=16$, and the one on the right was generated with $nheads=4$. Comparing both the heat maps, it can be seen that Fig. \ref{subfig:interaction1} has less number of active feature interactions per column, confirming our hypothesis that a larger number of attention heads leads to each head learning independent interactions between features.

\subsection{How does the Transformer scale with more data?}

To minimize the experimentation costs, all experiments in this paper were conducted using a training dataset of size 10 million. However, it is important to use the best performing model, the training dataset size can be increased to achieve optimal performance.

\begin{figure}
    \centering
    \includegraphics{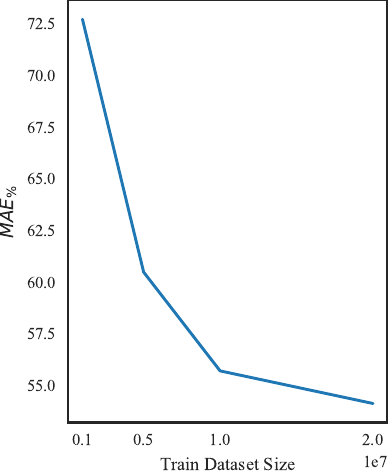}
    \caption{Scaling of RCT with data}
    \label{fig:scaling}
\end{figure}

To verify the scalability of RCT with data, we trained the model on different training dataset sizes and plotted the results in Fig. \ref{fig:scaling}. The results demonstrate that RCT's performance continues to improve with larger datasets. Therefore, we can confidently expect that models trained on larger datasets will outperform the model explored in this paper.

\section{Conclusion and Future Work}

In this paper, we presented a novel framework based on the Transformer architecture for predicting shipping costs on day 0. Our proposed framework encodes shipping attributes of a package, i.e., the package rate card, into a uniform embedding space. These embeddings are then fed through a Transformer layer, which models complex higher-order interactions and learns an effective representation of the package rate card for predicting shipping costs. Our experimental results demonstrate that the proposed model, called RCT, outperforms GBDT model by 28.8\%. Furthermore, demonstrate the RCT performs better than SOTA model FT-Transformer for our problem statement. We also show that when rate card representation learned by RCT is added to GBDT model, its performance is improved by 12.51\%. This underscores the fact that RCT is able to learn sufficient representation representations of rate card information.

In this work, the route information used was limited to the start and end nodes alone. Future work could explore the use of Graph Neural Networks to encode information about the complete route. Further, the performance of the RCT might be improved by exploring ways to include the item ID as a feature, such as the use of item embeddings which are available internally. 

Also, while the RCT was trained to predict only the ship cost, it can be modified to predict all the attributes of the invoice by adding a Transformer decoder layer. This would enable other applications like invoice anomaly detection. Additionally, future research could investigate whether the package representations learnt by the RCT can be used to improve the performance of other related tasks or to quantify the model uncertainty in each prediction via approaches like the one proposed in \cite{DER}.

\typeout{}  
\bibliography{src/tex/citation} 

\end{document}